\documentclass[letterpaper, 10 pt, conference]{ieeeconf}  

\IEEEoverridecommandlockouts                              

\title{\LARGE \bf
\fullname: Real-World Piano Playing via Fast Adaptation of \\ Dexterous Robot Policies
}

\author{Amber Xie$^{1}$, Haozhi Qi$^{2}$, Dorsa Sadigh$^{1}$
\thanks{$^{1}$Stanford University}
\thanks{$^{2}$Amazon FAR (Frontier AI \& Robotics)}
}

\usepackage{amsmath}
\usepackage{amssymb}
\usepackage{hyperref}
\usepackage{caption}

\usepackage{xcolor} 
\usepackage{graphicx} 
\usepackage{booktabs}
\usepackage{cleveref} 
\usepackage{multirow}
\usepackage{dsfont}
\usepackage{colortbl}
\usepackage{xspace}
\usepackage{bm}
\usepackage{cite}

\newcommand{\fullname}[0]{HandelBot\xspace}

\newcommand{\pisim}{$\pi_{sim}$\xspace}
\newcommand{\pisimbold}{$\bm{\pi_{sim}}$}
\newcommand\Std[1]{{\scriptsize \textcolor{gray}{$\pm$ #1}}}

\newcommand{\myparagraph}[1]{\vspace{0.2em}\noindent\textbf{#1}}

\begin{document}

\thispagestyle{empty}
\pagestyle{empty}

\let\oldtwocolumn\twocolumn
\renewcommand\twocolumn[1][]{%
\oldtwocolumn[{#1}{
\begin{center}
    \vspace{-1.5em}
    \includegraphics[width=\linewidth]{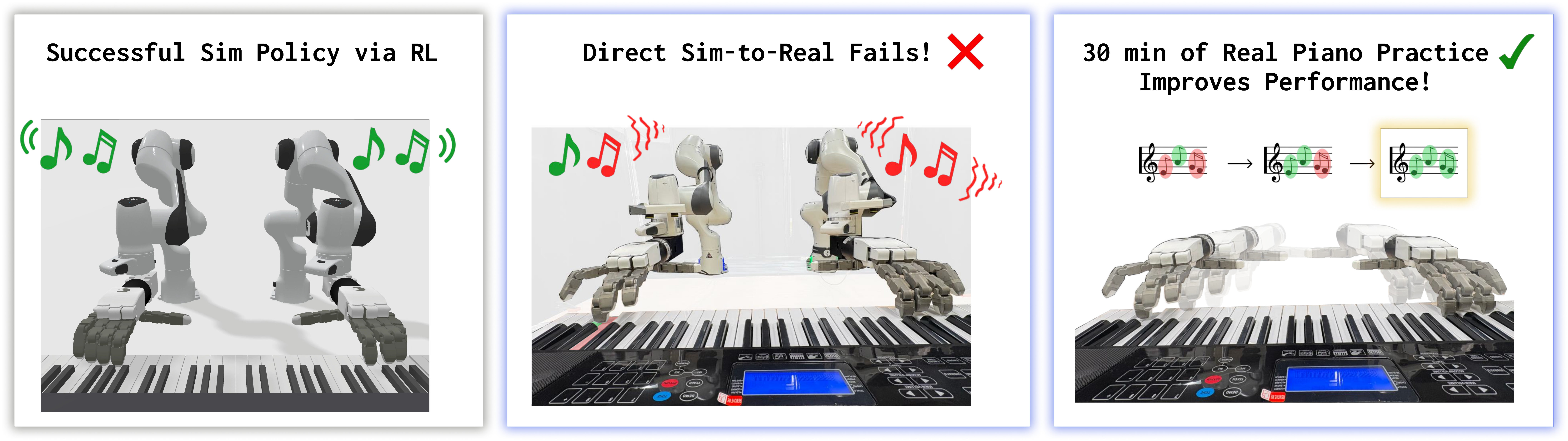}
    \captionof{figure}{We present \fullname, the first bimanual, dexterous piano-playing robot. For a spatially and temporally precise task like piano playing, directly deploying a simulation-trained policy fails, often hitting the wrong keys. Our key insight is to leverage real-world data to improve our simulation-trained policy. By applying structured trajectory refinement followed by residual reinforcement learning, \fullname achieves successful real-world execution in as little as 30 minutes.}
    \label{fig:teaser}
    \vspace{-0.2em}
\end{center}
}]
}

\maketitle

\begin{abstract}
Mastering dexterous manipulation with multi-fingered hands has been a grand challenge in robotics for decades. Despite its potential, the difficulty of collecting high-quality data remains a primary bottleneck for high-precision tasks. While reinforcement learning and simulation-to-real-world (sim-to-real) transfer offer a promising alternative, the transferred policies often fail for tasks requiring millimeter-scale precision, such as bimanual piano playing. In this work, we introduce \fullname, a framework that combines a simulation policy and rapid adaptation through a two-stage pipeline. Starting from a simulation-trained policy, we first apply a structured refinement stage to correct spatial alignments by adjusting lateral finger joints based on physical rollouts. Next, we use residual reinforcement learning to autonomously learn fine-grained corrective actions. Through extensive hardware experiments across five recognized songs, we demonstrate that \fullname can successfully perform precise bimanual piano playing. Our system outperforms direct simulation deployment by a factor of 1.8$\times$ and requires only 30 minutes of physical interaction data. Code and more videos are available at \url{https://amberxie88.github.io/handelbot}.

\end{abstract}

\section{INTRODUCTION}

Learning complex dexterous behaviors with multi-fingered hands remains a central challenge in robotics. Tasks such as piano playing represent a grand milestone in this domain, as they require precise, coordinated finger motions, fine contact timing, and long-horizon control. While recent advances have enabled impressive dexterous manipulation in simulation~\cite{robopianist2023}, real-world piano playing with robotic hands pushes far beyond what current systems can reliably demonstrate.

To tackle these complex dexterous tasks, imitation learning has emerged as a promising paradigm for achieving precise real-world control, but it relies heavily on large-scale, high-quality data~\cite{khazatsky2024droid, open_x_embodiment_rt_x_2023}. While teleoperation is a popular method for collecting this data, controlling high-DoF robot hands is cumbersome and lacks scalability. More importantly, it is often completely infeasible for tasks requiring rapid, independent finger motions, such as piano playing. Alternatively, learning directly from human data mitigates scalability issues but introduces a substantial embodiment gap, a problem heavily amplified by the spatial and temporal precision needed for this task. Reinforcement learning (RL) in simulation bypasses these issues entirely and has driven recent progress in simulated piano playing~\cite{robopianist2023}. However, relying solely on simulation introduces a sim-to-real gap that becomes highly problematic when millimeter-scale errors inevitably lead to task failure.

To address these challenges, we propose \textit{Hand}-elBot (inspired by Baroque composer George Frideric Handel), a reinforcement learning framework designed to bridge the gap between simulation and the real world. Our core insight is that while simulation struggles to model subtle contact dynamics, it is highly effective at providing a strong structural prior for finger coordination. Building on this, we decompose the learning process into a hybrid pipeline. We first utilize simulation to acquire a base policy of coarse motor movements. Next, we deploy this policy in the real world, refining it through structured lateral joint adjustments and residual reinforcement learning. By incorporating a small amount of high-quality physical interaction data, our policy efficiently adapts to the real world.

Specifically, we first train a policy in simulation. This achieves strong performance in the simulated environment but only gives limited success on real hardware due to the sim-to-real gap. To address this, we apply simple, human-defined heuristics that refine the physical trajectory to better match the expected performance. These heuristics exploit human priors of keyboard geometry and hand kinematics to correct consistent lateral biases and contact misalignments. By comparing the desired key presses to the actual keys pressed during a real-world rollout, we iteratively adjust the lateral joints of the fingers to align them horizontally with their targets. We subsequently refined policy via residual reinforcement learning, which learns corrective actions on top. This residual formulation enables safer exploration and effectively bridges the sim-to-real gap.

We extensively evaluate \fullname on a bimanual robot setup across a diverse suite of \textbf{five songs}. Our experiments demonstrate that directly deploying simulation policies struggles to accurately execute the tasks. In contrast, HandelBot consistently achieves the highest F1 scores across all evaluated musical pieces. We find that the initial policy refinement effectively aligns finger presses with target keys. Furthermore, the subsequent residual reinforcement learning stage significantly boosts performance by addressing errors and adapting to the physical dynamics.

To summarize, our contributions are as follows:
\begin{enumerate}
    \item To our knowledge, we present the first learning-based system capable of real-world, two-handed piano playing. We comprehensively evaluate this system across a diverse suite of five recognized songs.
    \item We propose a novel two-stage method for bridging the sim-to-real gap. This approach first refines a simulation-trained policy using physical rollouts, and subsequently utilizes real-world residual reinforcement learning to learn fine-grained corrective actions.
    \item We demonstrate that \fullname outperforms direct sim-to-real deployment by a factor of \textbf{1.8x}, with as little as 30 minutes of real-world interaction data.
\end{enumerate}

\section{RELATED WORK}

\myparagraph{Dexterous Manipulation.} Dexterous manipulation with multi-fingered robot hands offers the potential to achieve truly human-like motor skills. A promising approach to learning such skills is through imitation over teleoperated demonstrations~\cite{xu2025dexumi, zhang2025doglove,shaw2024bimanual,liu2019high,ding2025bunny,cheng2024open,iyer2024open,qin2023anyteleop,handa2020dexpilot,fang2025dexopdevicerobotictransfer}, similar to recent advances in robot learning for parallel jaw grippers~\cite{open_x_embodiment_rt_x_2023, zhao2023learningfinegrainedbimanualmanipulation,khazatsky2024droid,kim24openvla,gao2026taxonomy,@gao2024,intelligence2025pi05visionlanguageactionmodelopenworld,mirchandani2025robocrowdscalingrobotdata,mirchandani2025robocade,wu2023gello}. However, teleoperation is challenging for these high degree-of-freedom embodiments and lacks scalability. To tackle this, many works leverage direct human data~\cite{tao2025dexwild, guzey2025dexteritysmartlensesmultifingered, qin2022dexmv, kannan2023deft, wang2024dexcapscalableportablemocap, yin2025osmo, lum2025crossinghumanrobotembodimentgap}; yet, this introduces a substantial embodiment gap between human and robot morphology. An alternative paradigm trains policies entirely in simulation~\cite{openai2019solvingrubikscuberobot,yang2024anyrotategravityinvariantinhandobject,qi2022inhandobjectrotationrapid,kedia2026simtoolrealobjectcentricpolicyzeroshot,debakker2026scaffoldingdexterousmanipulationvisionlanguage,lum2024dextrahg}, which reduces the data collection burden but inevitably introduces a sim-to-real gap. 
When zero-shot sim-to-real does not lead to satisfactory task success, many methods leverage small amounts of real-world data for policy improvement~\cite{wang2024penspin,hsieh2025learning}.

\myparagraph{Robotic Piano Playing.} Early robotic piano systems relied on specialized hardware and handcrafted controllers designed explicitly for keyboard actuation~\cite{kato1987robot, lin2010electronic, topper2019piano, hughes2018anthropomorphic, castro2022robotic, zhang2009design, zhang2011musical, li2013controller, weng2025bidexhand}. Recent learning-based approaches have shown dexterous piano playing using general-purpose, non-specialized robotic hardware. These approaches leverage data-driven methods, including large-scale human motion capture \cite{wang2024furelise}, video-based kinematic retargeting \cite{qian2024pianomimelearninggeneralistdexterous}, and reinforcement learning in simulation \cite{xu2022learningplaypianodexterous,robopianist2023,zhao2024rp1m, chen2025dexterousroboticpianoplaying}.

Despite these advances, sim-to-real transfer for physical piano playing remains largely underexplored. While recent work~\cite{zeulner2025learningplaypianoreal} has demonstrated hybrid transfer for unimanual performance, we introduce a bimanual framework that couples simulation pretraining with real-world residual reinforcement learning, enabling complex, two-handed piano playing on non-specialized hardware.

\myparagraph{Real-World Reinforcement Learning.} Real-world reinforcement learning holds the potential of autonomous learning under real-world dynamics and environments. However, it remains challenging due to real-world reward functions, environment resets, safety and more. Prior work has shown that real-world RL from scratch is possible with careful design choices~\cite{smith2022walkparklearningwalk, hu2025robottrainsrobotautomatic, gupta2021resetfreereinforcementlearningmultitask, luo2025serlsoftwaresuitesampleefficient}. To address challenges with sample inefficiency and exploration in high-dimensional action spaces, prior works bootstrap with prior demonstrations~\cite{hu2023imitation, zhang2025rewindlanguageguidedrewardsteach,lei2026rl100performantroboticmanipulation}, multi-task or offline data~\cite{hu2023rebootreusedatabootstrapping, zhou2024efficient}, or large pretrained models~\cite{yang2023robotfinetuningeasypretraining,sobolmark2024policy}. In our work, we use an open-loop rollout learned from RL in sim, providing a strong initialization and reducing the exploration space.

Our specific direction of real-world RL focuses on residual reinforcement learning~\cite{johannink2018residualreinforcementlearningrobot}, which mitigates policy collapse and instabilities by keeping a frozen base policy. Prior work has explored residual policies for adaptation and exploration~\cite{yuan2024policydecoratormodelagnosticonline, ankile2025residualoffpolicyrlfinetuning, zhao2025resmimicgeneralmotiontracking}, generally assuming a strong pretrained policy that should be preserved. Our work is similar, as we learn a residual policy over an open-loop policy or trajectory. However, our approach does not assume access to expert demonstrations, which are difficult to obtain for piano playing, or closed-loop base policies. Our method enables efficient real-world learning to bridge the sim-to-real gap for a highly precise dexterous task.

\begin{figure*}[t]
\vspace{4mm}
    \centering
    \includegraphics[width=\linewidth]{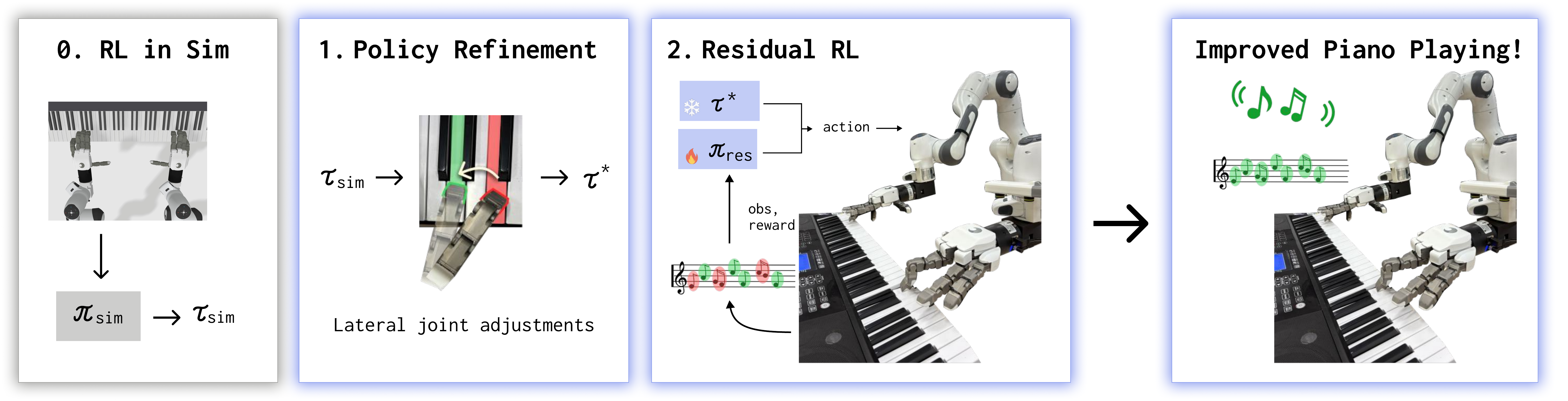}
    \caption{\textbf{\fullname Method} (0) \textbf{RL in Sim.} We leverage fast, parallel simulators for reinforcement learning. This leads to a coarse base policy, \pisim, from which we extract an open-loop rollout, $\tau_{sim}$. (1) \textbf{Policy Refinement.} Second, we refine $\tau_{sim}$, yielding $\tau^*_{sim}$. We use real-world updates to iteratively update the lateral joints of the fingers, moving the finger horizontally in the direction of the keys it is intended to press. (2) \textbf{Residual RL.} We perform residual RL atop $\tau_{sim}$, using the keyboard's MIDI output as a reward. This allows us to further update our policy for better piano playing.}
    \label{fig:method}
    \vspace{-1em}
\end{figure*}

\section{Real-World Piano Playing}
To adapt a simulation-trained policy for real-world piano playing, \fullname follows a two-phase process. First, we apply a structured policy refinement step to align the execution trajectory with the desired key presses, drawing inspiration from prior work~\cite{qian2024pianomimelearninggeneralistdexterous}. This alone still fails to reach the level of precision needed for piano playing. Second, \fullname uses residual reinforcement learning on top of the refined trajectory. The residual policy learns corrective action perturbations using real-world rewards. An overview of our approach is shown in Figure~\ref{fig:method}.

\subsection{Problem Statement}
We model robot piano playing as a Markov Decision Process (MDP), specified by the tuple $(\mathcal{O}, \mathcal{A}, P, r, \gamma)$. At each discrete time step $t$, the agent observes a state $o_t \in \mathcal{O}$, selects an action $a_t \in \mathcal{A}$ according to a policy $\pi(a_t \mid o_t)$, and transitions to a new state $o_{t+1} \sim P(\cdot \mid o_t, a_t)$ while receiving a scalar reward $r(o_t)$. We describe how we model robot piano playing as this MDP below in \cref{sec:rl}.

To learn our policy $\pi$, we use reinforcement learning, where $\pi$ is trained to maximize the expected discounted return:
\[
\mathbb{E}_{\pi}\left[ \sum_{t=0}^{T} \gamma^t r(o_t, a_t) \right],
\]
where $\gamma \in (0,1]$ is a discount factor and $T$ denotes the episode horizon.

\subsection{Reinforcement Learning in Simulation}
\label{sec:rl}
We first train a policy in a simulated environment to learn core piano-playing behaviors. We use RL, which leverages fast, parallel rollouts and dense reward signals, both of which are difficult to obtain in the real world.

\paragraph{Reward Design}
The simulated reward function largely follows the design of RoboPianist~\cite{robopianist2023}, and consists of a key press reward that rewards playing the target notes; a dense fingering reward for being near the correct key; and an energy penalty.

\paragraph{Observations and Actions}
Following prior work~\cite{robopianist2023}, the observation space includes robot proprioception, current piano activation, goal piano activations, and active fingers. The action space consists of delta joint positions, representing low-level control commands for the robotic hand. We script the end-effector based on physical location of keys to be pressed. We denote the resulting simulation policy as $\pi_{\text{sim}}$.


While $\pi_{\text{sim}}$ achieves strong performance in simulation, direct deployment in the real world leads to degraded performance due to mismatches in the controller and piano key-pressing dynamics.

\subsection{Policy Refinement}
Before running residual reinforcement learning, we first apply a lightweight policy refinement procedure in the real world. This is done by exploiting domain knowledge about keyboard geometry and hand kinematics to correct consistent lateral biases and contact misalignments. This can be viewed as a structured, system-identification-inspired trajectory refinement procedure. Empirically, this stage substantially improves key accuracy and provides a stronger initialization for subsequent real-world residual reinforcement learning.

The goal of this step is to iteratively correct systematic key-pressing errors in an \textit{open-loop rollout} of $\pi_{\text{sim}}$ without requiring additional learning. We denote the initial open-loop trajectory from $\pi_{\text{sim}}$ as $\tau^0 = (s_0^0, ..., s_T^0)$, where $s$ are the target joint states for timesteps $0, ..., T$. Concretely, we would like to produce $\tau^* = (s_0^*, ..., s_T^*)$, where $\tau^*$ denotes our corrected, improved trajectory.

\paragraph{Lateral Joint Correction}
In our embodiment, each finger is controlled by 4 joints, with a primary lateral joint responsible for horizontal motion and 3 others used for vertical actuation. After executing $\tau^0$ on the real piano, we compare desired keys presses to actual key presses, and use this information to adjust the lateral joint of the finger. For example, if the finger plays a lower-pitched note than the desired key, the finger should be adjusted to the right.

Specifically, we first execute the simulation-trained policy $\pi_{\text{sim}}$ in an open-loop manner on the real robot and record, at each time step $t$, (i) the target note and finger designated to press the key, and (ii) the set of keys actually pressed by each finger. Let $k_t^{\text{target}}$ denote the intended key index for a given finger, and let $\mathcal{K}_t^{\text{press}}$ denote the indices of keys that were physically activated.

For each finger, we identify the pressed key closest to the target. If the closest key $k_t^{\text{press}}$ differs from $k_t^{\text{target}}$, we compute a signed directional error:
$$
\Delta_t =
\begin{cases}
+\delta & \text{if } k_t^{\text{press}} < k_t^{\text{target}}, \\
-\delta & \text{if } k_t^{\text{press}} > k_t^{\text{target}}, \\
0 & \text{otherwise},
\end{cases}
$$
where $\delta$ is a step size controlling the amount of adjustment to the lateral finger joint. Then, we apply an update function $f : (\tau_0, \Delta_t) \mapsto \tau_1$ to obtain our updated trajectory, which we describe below in Chunked Updates.

\paragraph{Iterative Updates}
We apply this correction procedure iteratively, where we alternate between executing the current trajectory and updating it. During these iterations, we must determine how much to update the lateral joint, i.e. the value of $\delta$. We initialize $\delta$ to a large value, and after every iteration, we anneal $\delta$, to help avoid oscillation and allow for smoother convergence to an improved trajectory.

In practice, there are a few additional details that improve this process. First, we also introduce smaller corrective terms, $0.3\Delta_t$, to neighboring fingers, encouraging spatial separation between adjacent fingers and reducing self-collisions. Second, there may be many $k_t^{target}$ at each step, and therefore many $\Delta_t$ at each timestep $t$. For simplicity of notation, we previously defined $f$ under the assumption that there is only one key press and one potential lateral movement at each timestep. In reality, we calculate $\Delta_t$ for each active key press, and apply the correction to each lateral joint. Finally, when there are multiple key presses, we must also determine which key is pressed by which finger. In this case, we assume, based on finger and piano kinematics, that within $\mathcal{K}_t^{\text{press}}$, the active finger to the left is pressing the lower keys, and the active finger to the right is pressing the higher keys.

\paragraph{Chunked Updates}
Previously, we described our update function $f : (\tau_0, \Delta_t) \mapsto \tau_1$ as operating over each state in the open-loop rollout. However, in practice, we perform updates over temporal chunks for motion smoothness. Intuitively, this allows us to incorporate the temporal context of a segment; for instance, if a finger must move rightward to strike a key, the corrective residual should initiate a preparatory lateral movement during the approach phase, rather than a discrete correction at the moment of contact. 

Specifically, we divide the trajectory into sub-chunks of length $K$. Instead of computing $\Delta_{t}$ for each $s_t^i$ at iteration $i$, we instead compute  $\Delta_{t}^{\text{chunk}}$ for each chunk $s_t^i, ..., s_{t+K}^i$. To do so, we calculate $\Delta_{t}, ..., \Delta_{t+K+L}$, with lookahead $L$. Notice here that for a chunk from $t$ to $t+K$, we consider fingertip errors all the way to $t+K+L$, facilitating anticipatory spatial adjustments. To extract our $\Delta_{t}^{\text{chunk}}$ which we will apply to our entire chunk, we use $\Delta_{t}^{\text{chunk}} = \frac{1}{K+L} \sum_{j=t}^{t+k+L} \Delta_{j}$. 

At the end of this iterative process, we save the trajectory with the best F1 score as our refined trajectory.

\subsection{Real-World Residual Reinforcement Learning}
To address the sim-to-real gap, we adopt a residual reinforcement learning framework that fine-tunes the open-loop trajectory $s^*_0, ..., s^*_T$ from the policy refinement step.
\paragraph{Residual Policy Formulation}
We introduce a residual policy $\pi_{\text{res}}$ that outputs an additive correction to the base action:
\[
\hat{s}_{t+1} = \pi_{\text{res}}(o_t) + s^*_{t+1},
\]
where $o_t$ denotes the real-world observation at time $t$, $s^*_{t+1}$ denotes the next state from the open-loop trajectory, and $\hat{s}_{t+1}$ incorporates the residual action. Note here that $\hat{s}_{t+1}$ is roughly equivalent to $a_t$: because we are commanding absolute joint positions of the hands, our actions are simply the next target joint states. 

The residual action space is intentionally constrained to small perturbations, enabling safer exploration and faster learning compared to learning a full policy from scratch.

\paragraph{Residual RL Objective}
In the real world, we rely exclusively a reward signal derived from the piano's MIDI output. Our reward is simply the key press reward, identical to the one used in simulation. 

The residual policy $\pi_{\text{res}}$ is trained using reinforcement learning to maximize the expected return under the real-world dynamics. By learning only corrective action deltas, residual reinforcement learning effectively compensates for simulation inaccuracies while maintaining the structured behavior learned in simulation. This approach enables efficient and stable adaptation, resulting in robust real-world piano-playing performance.

\begin{figure}
    \centering
    \vspace{2mm}
    \includegraphics[width=1.0\linewidth]{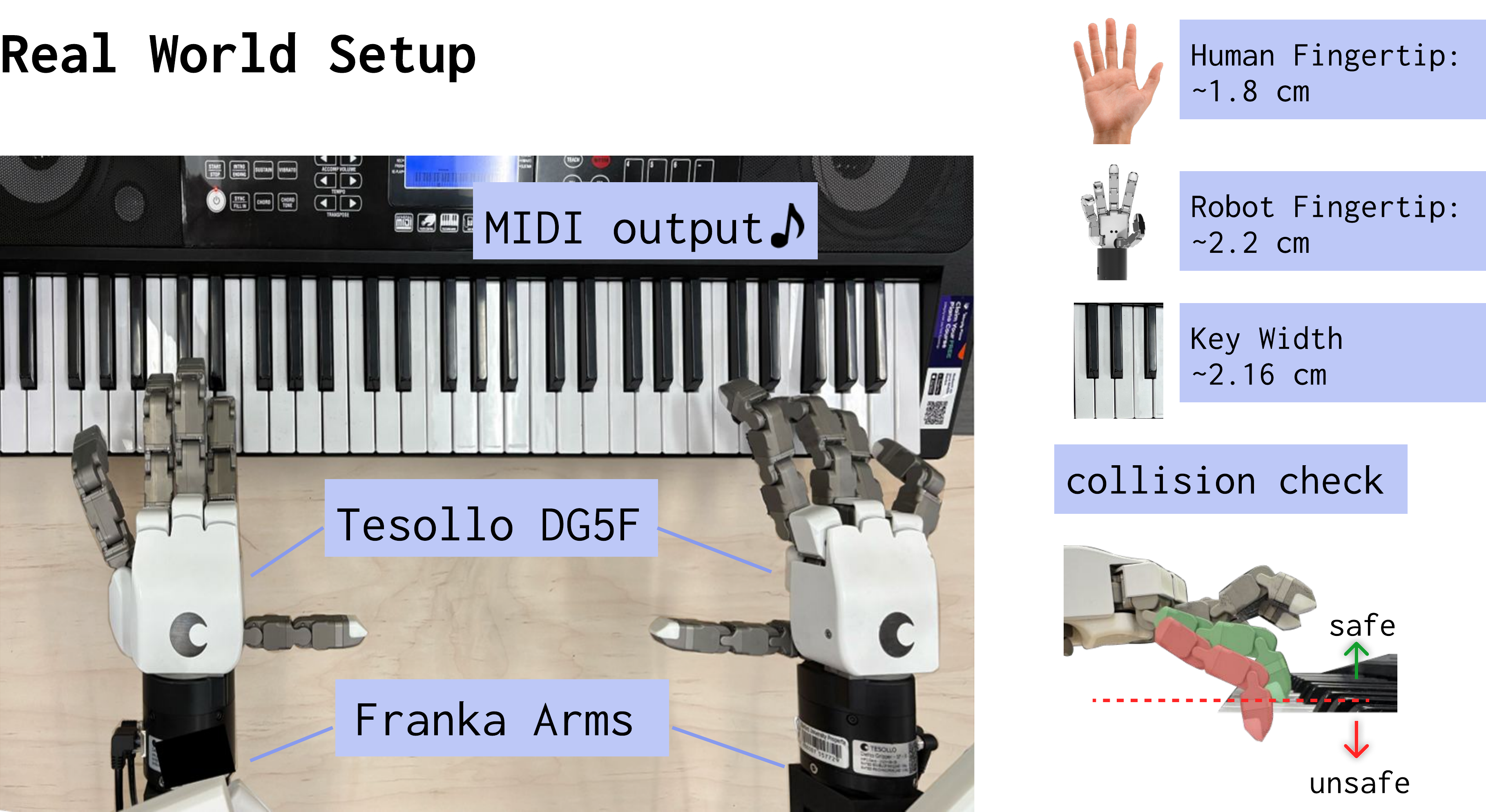}
    \caption{\textbf{Hardware Setup}. We use a MIDI keyboard, two Tesollo DG-5F hands, and two Franka arms for piano playing. We use the MIDI output from the piano, which tells us which notes are pressed, in order to calculate rewards. We emphasize that the robot hands are far larger than the average human hand, thus making piano playing difficult. Finally, for RL training, we include a collision checker which prevents fingers from pressing down beyond the keys.}
    \vspace{-2em}
    \label{fig:hardware}
\end{figure}

\begin{figure*}[t]
\vspace{4mm}
    \centering
    \includegraphics[width=\linewidth]{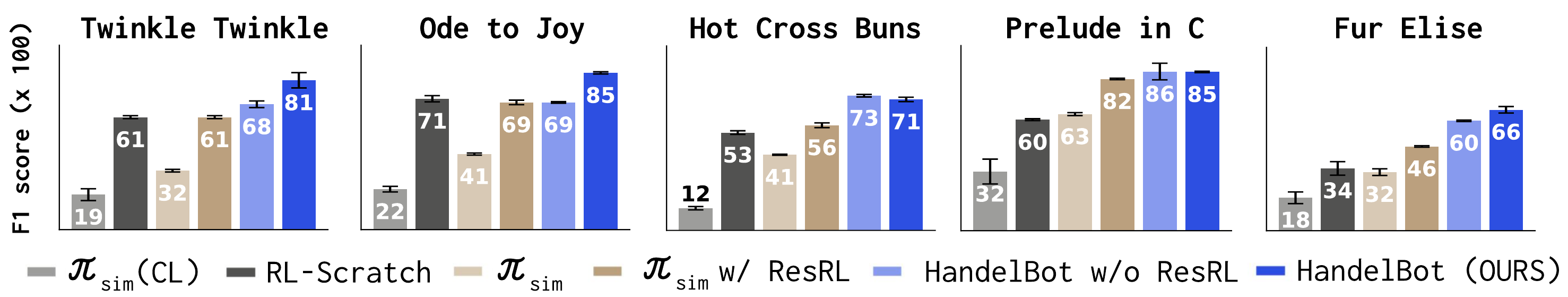}
    \caption{\textbf{Main Results.} We include F1 score, multiplied by 100, for 5 songs. \fullname consistently achieves the strongest F1 score, showing the importance of effectively using real-world samples to accomplish precise, dexteorus piano-playing. Methods only using simulated data, such as \pisim (CL) and \pisim, have weak performance due to the sim-to-real gap.}
    \label{fig:main}
\end{figure*}

\paragraph{Guided Noise} 
\label{sec:gnoise}
We choose TD3~\cite{fujimoto2018addressing} as our RL algorithm, which adds a noising term to the sampled action from the actor. Concretely, for each environment step, the actor computes $a = \mu_\theta(o) + \text{clip}( \epsilon, -0.5, 0.5)$, where $
\epsilon \sim N(0, 1)$. Motivated by the lateral adjustment used in policy refinement, we also adjust the noise in the direction of the correct lateral movement. We thus apply $a = \mu_\theta(o) + \text{clip}( \hat{\epsilon}, -0.5, 0.5)$, where $\hat{\epsilon}$ is a modification of $\epsilon$ as follows:

First, for any keys pressed, we calculate the signed directional error $\Delta_t$ for the current timestep, as we did in policy refinement. With probability $\Pr(\text{guided noise}) = 0.5$, we change the sign of the noise at that lateral joint to be the same sign as $\Delta_t$. This produces $\hat{\epsilon}$, where $||\hat{\epsilon}||_2 = ||\epsilon||_2$. Note that our only modification is the sign of the noise term for the action indices corresponding to the lateral joints. Then, if guided noise is sampled, the final action our actor takes is $a = \mu_\theta(o) + \text{clip}( \hat{\epsilon}, -0.5, 0.5)$.

This is a lightweight heuristic that guides exploration to hitting the correct keys. In practice, we do not find this to be an important hyperparameter, as explored in \cref{tab:abl}.

\section{Experiments}

\subsection{Experimental Setup}

\subsubsection{Hardware Platform}

Our physical system consists of two Tesollo DG-5F hands mounted on a Franka Emika Panda arm and a FR3 arm. To better emulate human piano-playing posture, we design a custom 3D-printed mount so the the fingers are parallel to the keyboard surface. The arms follow a script end-effector pose trajectories, and the hands are controlled with joint control.

\subsubsection{Simulation Environment}
We use ManiSkill~\cite{gu2023maniskill2}, a fast, parallelizable simulator, for reinforcement learning in simulation. For each trained policy, we select the trajectory achieving the highest validation F1 score and treat it as the nominal open-loop solution for real-world deployment. This selection protocol ensures that sim-to-real transfer begins from the strongest available simulation behavior.

\subsubsection{Real-World Deployment and Safety}

Direct execution of simulated trajectories on hardware introduces safety risks and control instability. We therefore augment deployment with a real-world safety layer. Given desired joint targets, we solve a constrained inverse kinematics problem using PyRoki~\cite{pyroki2025} to produce feasible configurations while penalizing self-collisions and contact with the piano surface, which we approximate as a planar constraint. To improve motion smoothness, policy actions are produced at 10Hz and linearly interpolated to 80Hz before being sent to the hands. This upsampling reduces jerk and high-frequency oscillations that are particularly detrimental in piano playing. The arms are controlled using the Polymetis controller at 100Hz to ensure stable end-effector tracking.

\begin{figure*}[!t]
\vspace{4mm}
    \centering
    \includegraphics[width=\linewidth]{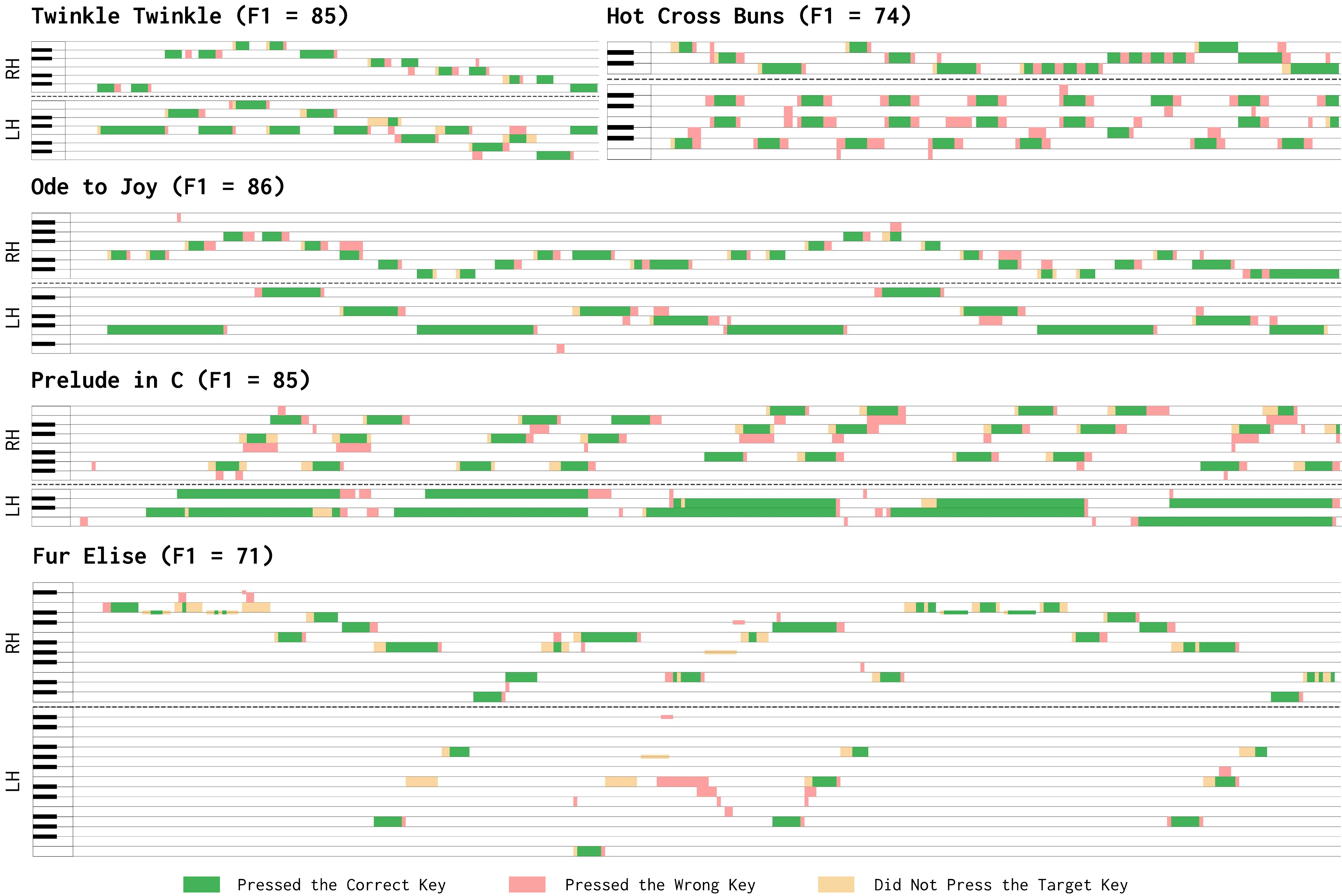}
    \caption{\textbf{Visualization of \fullname Trajectories.} Per each song, we visualize the notes pressed \textcolor[HTML]{149F2E}{correctly}, pressed \textcolor[HTML]{FF2525}{incorrectly}, and \textcolor[HTML]{F39C12}{missed}. The x axis is the timestep of the song, and the y axis are the different notes, with the top half representing keys for the right hand, and the bottom for the left hand. Across easier songs such as Twinkle Twinkle and Ode to Joy, we find that \fullname makes few mistakes, with occasional timing errors or wrong presses. For harder songs such as Fur Elise, large jumps in the left hand notes (bottom section of each song plot) are challenging for the left hand.}
    \label{fig:vis}
\end{figure*}

For residual reinforcement learning, we restrict the residual actions to only the three active fingers (index, middle, ring), with total dimensionality 9 per hand. Each hand is trained as an independent agent, which reduces action dimensionality and simplifies credit assignment during adaptation. 

\subsubsection{Musical Tasks}
We evaluate on 5 widely recognized songs with varying song lengths: Twinkle Twinkle (160 timesteps; 16 seconds), Ode to Joy (330 timesteps; 33 sec), Hot Cross Buns (160 timesteps; 16 sec), Fur Elise by Beethoven (320 timesteps; 32 sec), Prelude in C by Bach (330 timesteps; 33 sec). Due to physical reach and lateral dexterity constraints for the thumb and pinky, we modify the fingerings for the song to three fingers per hand. To prevent inter-arm collisions, we arrange left and right hand parts to be separated by multiple octaves. 

\subsubsection{Evaluation Protocol}
Following prior work~\cite{robopianist2023}, we measure policy performance using the F1 score. For all methods, we report the mean F1 score across 5 rollouts. For reinforcement learning methods, policies are evaluated with the mean of 5 rollouts, after every 20 trajectories during training, and we report the maximum validation F1 score achieved over the course of learning. RL methods are trained for 100 trajectories. This roughly corresponds to 30k environment interactions in 1 hour for Ode to Joy, Prelude in C, and Fur Elise; and roughly 16k environment interactions in 30 minutes for Twinkle Twinkle and Hot Cross Buns. 

\subsubsection{Baselines}
We compare against a comprehensive set of baselines to isolate the contribution of each component in our method. We use TD3 for all RL methods~\cite{fujimoto2018addressing}.

\begin{itemize}
\item \textbf{\pisimbold (CL)}: policy trained in simulation with domain randomization, with closed-loop inference.

\item \textbf{RL from Scratch}: a real-world policy from random initialization, using scripted end-effector motion but learning finger control entirely on hardware with real-world dynamics.

\item \textbf{\pisimbold}: open-loop policy from sim.

\item \textbf{\pisimbold + ResRL}: residual RL over $\pi_{sim}$.

\item \textbf{\fullname w/o ResRL}: policy refinement over $\pi_{sim}$.

\item \textbf{\fullname (Ours)}: policy refinement and residual reinforcement learning, yielding our full system.
\end{itemize}

\subsection{Can \fullname Accomplish Real-World Piano Playing?}
In \cref{fig:main}, we examine the performance across a suite of 5 songs. First, we find that real data consistently boosts performance of piano-playing. For example, real-world RL from scratch, which leverages no prior pretraining, is able to achieve strong performance for many songs, matching or outperforming \pisim. Applying residual RL over \pisim also leads to consistent improvements. In contrast, methods using no real data, like \pisim (CL) and \pisim, have the lowest F1 scores. Notably, \pisim (CL) performs much worse than open-loop \pisim, which we hypothesize is because of the dynamics gap and compounding errors across the trajectory. Qualitatively, we find that \pisim and \pisim (CL) often overshoot past the target key, and often may press down forcefully and get stuck on a key, issues that using real data addresses.

\begin{table}[]
\caption{We report F1 scores x 100 for the closed-loop \pisim, which runs inference over real-world inputs, with hybrid execution \pisim. Hybrid execution runs a simulation in parallel with the real-world, and uses proprioception from the parallel sim instead of the real-world.}
\centering
\label{tab:sim2real}
\begin{tabular}{l|ll}
\toprule
\textbf{Song} & \pisimbold (\textbf{CL}) & \pisimbold (\textbf{Hybrid}) \\
\midrule
Ode to Joy & 5 \Std{2.46} & 12 \Std{2.6}  \\
Twinkle Twinkle & 23 \Std{6.2} & 24 \Std{2.5} \\
Hot Cross Buns & 8 \Std{2.1} & 9 \Std{2.9} \\
Prelude in C & 29 \Std{2.2} & 40 \Std{1.0} \\
Fur Elise & 18 \Std{3.2} & 20 \Std{4.9} \\
\bottomrule
\end{tabular}
\end{table}

\begin{table}[]
\caption{\textbf{Ablations}. By default, \fullname uses $\gamma = 0.8$ and $\Pr(\text{guided noise}) = 0.5$.}
\centering
\label{tab:abl}
\begin{tabular}{l|l}
\toprule
 & \textbf{Twinkle} \\
\midrule
\fullname (OURS) & 81 \Std{4.1} \\ 
\fullname: $\gamma$ = 0.75 & 73 \Std{2.5} \\
\fullname: $\gamma$ = 0.9 & 69 \Std{0.2} \\
\fullname: $\Pr(\text{guided noise})$ = 0 & 81 \Std{0.7} \\
\fullname: $\Pr(\text{guided noise})$ = 1 & 77 \Std{0.9} \\

\bottomrule
\end{tabular}
\end{table}

Across all songs, \fullname is the strongest performing method, leveraging both policy refinement and residual RL for enhanced performance. We find policy refinement to be effective, as it directly aligns finger presses with the correct target keys. However, policy refinement cannot address many issues, which is why residual RL is still necessary. First, we only adjust the lateral joint during refinement, meaning that missed presses cannot be adjusted. Second, refinement makes a series of assumptions, such as which finger is pressing which key, and this may not always be accurate. This is addressed by learning the end-to-end residual policy, which leads to a boost in performance across most songs. 

\subsection{What Factors are Important for Real-World Residual RL?}
Next, we examine the effectiveness of real-world RL. First, we evaluate how initializing RL with a pretrained trajectory impacts performance. We include 3 variants: RL-Scratch, which learns the entire hand policy without prior pretraining; \pisim + ResRL, which learns residuals over \pisim; and \fullname, which learns residuals over a refined trajectory. Across all songs, we find that residual RL over a strong trajectory (\pisim with policy refinement $>$ \pisim $>$ no initialization) leads to higher F1 scores. We hypothesize that a refined policy reduces the exploration space, leading to stabler and more efficient training.

Next, we ablate RL design decisions in \cref{tab:abl}. Following prior piano work~\cite{robopianist2023}, we try 2 values of $\gamma$, the RL discount factor, and find that a lower discount leads to lower F1 scores, and qualitatively, jerkier motions. We ablate the probability of sampling guided noise, and find our default hyperparameter is similar to not sampling guided noise. However, always sampling guided noise leads to degraded performance, which we hypothesize is because finger exploration is biased, which prevents learning from suboptimal data. 

\begin{figure}
    \centering
    \vspace{2mm}
    \includegraphics[width=\linewidth]{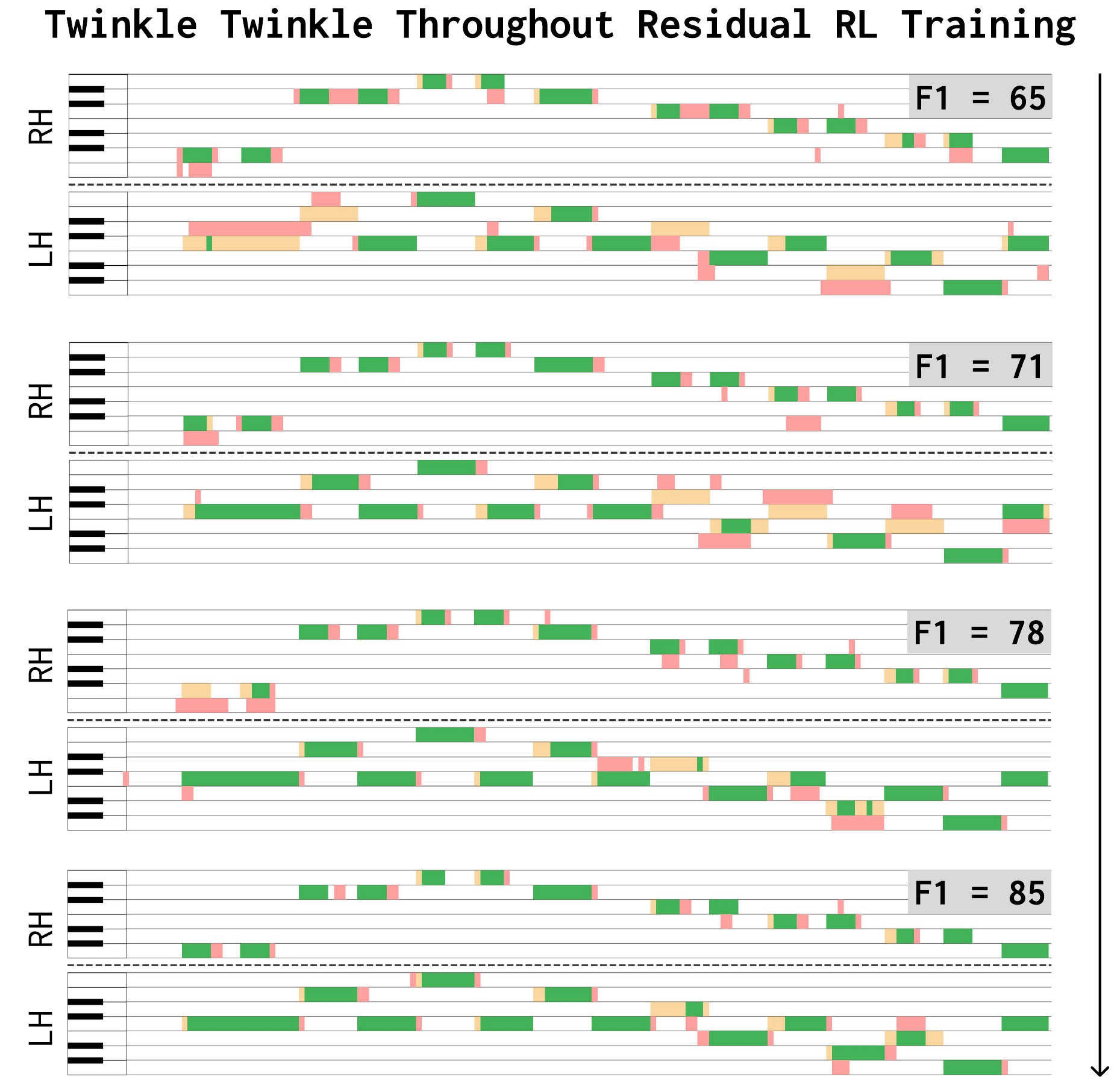}
    \caption{\textbf{\fullname Trajectories across Residual RL Training.} We include 4 evaluation trajectories during \fullname training, with the final, best-performing trajectory in \cref{fig:vis}. Across these 4 trajectories, we see that \fullname initially struggles with many keys in the left hand. However, with real-world interactions, the residual policy is able to adapt to real world and press the correct keys.}
    \label{fig:twinkle}
\end{figure}

\subsection{Ablation: Closed-Loop Sim-to-Real}
In \cref{tab:sim2real}, we compare closed-loop inference and hybrid execution~\cite{zeulner2025learningplaypianoreal} for closed-loop sim-to-real policies. Hybrid execution, an alternative to direct sim-to-real transfer, runs a simulated environment in parallel with the real environment. Instead of taking real-world observations, hybrid execution mitigates the sim-to-real gap by using simulated observations. Similar to \cite{zeulner2025learningplaypianoreal}, we find an improvement when using hybrid execution, but performance is still far from \fullname and other methods that utilize real-world data. We believe this is because hybrid execution only artificially reduces the gap between sim and real, and it cannot properly adjust to real-world dynamics.

\section{CONCLUSIONS}

In this paper, we introduced \fullname, which leverages reinforcement learning in simulation, policy refinement, and residual reinforcement learning to tackle the extreme precision requirements of robotic piano playing. Our key insight is treating simulation as a foundation for global motor coordination and real-world interaction as the mechanism for fine-grained refinement. Our results demonstrate that our method can transform a brittle, imperfect simulated policy into a much more robust piano playing robot, requiring as little as 30 minutes of real-world data. We believe \fullname represents a significant step toward deploying high-DoF dexterous hands in environments where spatial and temporal timing is the difference between success and failure. 

\myparagraph{Limitations.} \fullname  relies on scripted end-effector movements with a fixed orientation, which leads to various limitations. First, this requires some amount of manual tuning each time. Residual RL over the end effector movements may reduce this problem. Second, this makes using the thumb and pinky more difficult. For this reason, we only evaluate on relatively simple songs. Future work may explore allowing rotations or learned movements in order to better utilize other fingers for more complex songs. Another limitation is the policy refinement step, which depends on human-guided heuristics. While this is sensible for piano playing and exploits obvious domain knowledge, this is not directly applicable to other tasks. However, policy refinement is possible for other tasks, either by human-guided heuristics or large models like vision-language models, which can help refine the policy at a coarse level before residual RL. 

\section*{ACKNOWLEDGMENT}
This work was funded by ONR MURI N00014-25-1-2479, and ONR YIP N00014-22-1-2293, NSF \#2218760 and NSF \#1941722. We would like to thank members of ILIAD for their feedback and support. We thank Satvik Sharma and Jayson Meribe for hardware assistance and support, Marcel Torne Villasevil and Megha Srivastava for project brainstorming and simulator development, Hung-Chieh Fang for hardware and software advice, and Abrar Anwar and Hengyuan Hu for RL advice. 

\bibliographystyle{IEEEtran}
\bibliography{IEEEabrv,full_references}

\clearpage

\section*{APPENDIX}
We open-source our simulated and real-world implementations in \url{https://github.com/amberxie88/handelbot} and show videos on our website \url{https://amberxie88.github.io/handelbot}.

\subsection{Simulation Training}
We train a PPO~\cite{schulman2017proximal} policy per song. However, instead of a power penalty, we instead penalize an action L1 penalty on the delta finger joint actions. We also ablate the Key Press reward with a different Key On coefficient. The resulting key press reward, copying RoboPianist notation, is: $0.7 \cdot \left( \frac{1}{K} \sum_{i}^K g(||k_s^i - 1 ||_2) \right) + 0.3 \cdot (1 - \mathbf{1}_{\{\text{false positive\}}})$, where $K$ is the number of keys that need to be pressed at the current timestep, $k_s$ is the normalized joint position of the key between 0 and 1, and $\mathbf{1}_{\{\text{false positive\}}}$ is an indicator function that is 1 if any key that should not be pressed created a sound~\cite{robopianist2023}. We find this necessary to encourage key presses at the expense of some wrong key presses. For this embodiment, wrong key presses, even in simulation, are almost unavoidable, hence this modification.

Furthermore, we fix the last joint angle of all fingers to be 1 radian. This reduces the action space to 3 joints per finger, reducing RL exploration and optimization. We found this helpful because the Tesollo DG-5F fingers are quite large, but the fingertips are slightly narrower. By fixing the last joint of each finger, the finger presses the key with only the tip of the finger, leading to a more human-like position and potentially a lower likelihood of pressing multiple keys.

To script the end effector position, we calculate the wrist trajectory based on the sheet music. The wrist trajectory is derived from the note sequence by computing the wrist position required to place the specified finger on each target key. For every note, the Y position of the key is obtained from the piano geometry, and a predefined finger–wrist offset is subtracted so that the corresponding finger aligns with the key when the wrist is at that position. The X position is set based on whether the key is white or black, since black keys are located slightly farther away.

When multiple notes occur at the same timestep, the wrist targets are aggregated by averaging the Y positions and selecting the minimum X position. These positions define anchor points in time, and a continuous wrist trajectory is then produced by linearly interpolating between them across all timesteps. Finally, the trajectory is expressed relative to the robot’s initial wrist pose.

Additional hyperparameters are described in \cref{tab:simhps}.

\begin{table}[h]
\caption{\textbf{Hyperparameters for Simulation RL}.}
\centering
\label{tab:simhps}
\begin{tabular}{l|l}
\toprule
Total Timesteps & 40,000,000 \\
\# Minibatches & 32 \\
Update Epochs & 8 \\
\# Steps & 32 \\
Fingering Rew. Coef. & 1 \\
Key Press Rew. Coef. & 1 \\
Action L1 Rew. Coef & 0.01 \\
Key On Coef. & 0.7 \\
Gamma & 0.8 \\
\bottomrule
\end{tabular}
\end{table}

\subsection{Real-World Training}
For real-world residual RL, we run an actor and learner process separately, as inspired by~\cite{luo2025serlsoftwaresuitesampleefficient}. Our residual policy has chunk size 2, meaning that each residual action the policy outputs is repeated twice, effectively learning residuals at 5Hz over the 10Hz simulated trajectory. We find this leads to smoother learning. Next, we also sample correlated noise for our TD3 agent to reduce jitter. Concretely, our noise sampled at every step is $\hat{\epsilon} = \beta * \epsilon_{prev} + \sqrt{1 - \beta^2} * \epsilon$, where our correlation coefficient $\beta = 0.2$, $\epsilon \sim N(0, 1)$, and $\hat{\epsilon}$ is the correlated noise we actually use. We also use a linear noising schedule for the first 10,000 gradient steps of policy training.

We include hyperparameters in \cref{tab:realhps}.

\begin{table}[h]
\caption{\textbf{Hyperparameters for Real-World RL}.}
\centering
\label{tab:realhps}
\begin{tabular}{l|l}
\toprule
Total Timesteps & 100 trajectories \\
\# Critics & 2 \\
Guided Noise Probability & 0.5 \\
Batch Size & 2048 \\
$\tau$ for target critic & 0.005 \\ 
\# Initial Exploration Steps & 512 \\
Learning Rate & 1e-3 \\
Dropout & 0.5 \\
Actor / Critic Hidden Dimensions & [256, 256, 256] \\
Update to Data Ratio & 8 \\
Update Actor Every / Update Critic Every & 10 \\
Initial Key On Coef. & 0.7 for 10,000 grad steps \\
Final Key On Coef. & 0.5 for remaining \\
Gamma & 0.8 \\
\bottomrule
\end{tabular}
\end{table}

\subsection{Piano Songs}
\begin{enumerate}
\item \textbf{Twinkle Twinkle.} We adapt Twinkle Twinkle from the original RoboPianist~\cite{robopianist2023} MIDI file, making minor modifications such as changing the octaves for each hand and reannotating fingerings.
\item \textbf{Ode to Joy.} We use a popular YouTube version with minor modifications: \url{https://www.youtube.com/watch?v=i4bfhW7uLmc}
\item \textbf{Hot Cross Buns.} We compose a simple version of the popular nursery rhyme.
\item \textbf{Prelude in C.} We change octaves and annotate fingerings for the original classical piece, Prelude and Fugue in C major, BWV 846 by Johann Sebastian Bach.
\item \textbf{Fur Elise.} We change octaves for the original classical piece, Für Elise by Ludwig Van Beethoven.
\end{enumerate}

\end{document}